# How To Grade a Test Without Knowing the Answers — A Bayesian Graphical Model for Adaptive Crowdsourcing and Aptitude Testing

**Yoram Bachrach**  YOBACH@MICROSOFT.COM
Microsoft Research, Cambridge, UK

**Tom Minka**  MINKA@MICROSOFT.COM
Microsoft Research, Cambridge, UK

**John Guiver**  JOGUIVER@MICROSOFT.COM
Microsoft Research, Cambridge, UK

**Thore Graepel**  THOREG@MICROSOFT.COM
Microsoft Research, Cambridge, UK

## Abstract

We propose a new probabilistic graphical model that jointly models the difficulties of questions, the abilities of participants and the correct answers to questions in aptitude testing and crowdsourcing settings. We devise an active learning/adaptive testing scheme based on a greedy minimization of expected model entropy, which allows a more efficient resource allocation by dynamically choosing the next question to be asked based on the previous responses. We present experimental results that confirm the ability of our model to infer the required parameters and demonstrate that the adaptive testing scheme requires fewer questions to obtain the same accuracy as a static test scenario.

## 1. Introduction

Collective decision making is a well-studied topic in social choice, voting and artificial intelligence. It has long been known that decisions based on aggregating the opinions of several agents can be of higher quality than those based on the opinions of single individuals. The Condorcet Jury Theorem (de Caritat et al., 1785), dating back to the 18th century, is concerned with a group of individuals attempting to reach a binary decision by majority vote; it is assumed that one of the two outcomes of the vote is "correct", and that each individual independently chooses the "correct" response with probability $p$. The theorem states that if $p > \frac{1}{2}$, then adding more agents increases the probability of making the correct decision, and the probability that the collective decision will be "correct" approaches 1 in the limit of infinitely many participating agents.

Recent technological advances make it easier to share opinions and knowledge, enabling us to harness the collective intelligence of crowds for solving tasks. Companies can use *crowdsourcing* to carry out business tasks, using platforms such as Amazon Mechanical Turk. Such services allow us to collect the opinions of many individuals, but leave open the question of how to aggregate the collected data to reach decisions.

A key technique for solving tasks using collective intelligence is to obtain information from multiple sources and aggregate it into a single complete solution. Consider a crowd of experts who are assigned with many similar classification tasks, such as classifying many news articles according to their topic ("politics", "business", "entertainment" etc.). We refer to each expert as a *participant* and to each classification task as a *question*. Suppose each participant expresses her opinion regarding the correct answer for each question, in the form of a *response*, chosen by her from the list of possible answers for that question. Similarly to Condorcet's Jury Theorem, we make the simplifying assumption that for each of the questions only one answer is *correct*. We call such a domain a *multiple problem domain*. Given the responses provided by the participants regarding the various questions in a mul-tiple problem domain, how should we best determine the correct answer for each of the items? Which questions are easy and which are hard? How can we find the most competent participants in the crowd? Which questions best test the ability of a participant?

Given the correct answers to each of the items, it is easy to find the most competent participants, or differentiate the easy and hard questions — a participant who has answered almost all the questions correctly is likely to be more skilled than a participant who had very few correct responses. Typically, however, the correct answers are not known in advance — the whole point of crowdsourcing a classification task is to determine the correct classifications for the items.

A possible solution to the above problem is to first evaluate the skill of each expert by asking her to provide responses to a set of items for which the correct answer *is known* (sometimes called a "gold-set"). A prominent example of this, where *all* correct answers are known, is intelligence testing (Anastasi et al., 1997). Psychologists have studied human intelligence, and designed IQ tests for evaluating the aptitude of individuals. These tests have been shown to be predictive of a person's performance in many domains, such as academic attainment and job success (Anastasi et al., 1997). Such tests typically ask participants to respond to questionnaires composed of many multiple-choice questions, and allow ranking participants according to their individual skill levels after examining the responses. The properties of responses to IQ tests have been widely studied by psychometricians, so such datasets can serve as a testing ground for exploring inference models for multiple problem domains.

**Our Contribution:** We propose a new family of graphical models for analyzing responses in multiple problem domains and evaluate the models on a data set of completed questionnaires of a standard IQ test. The proposed framework enables us to jointly infer the correct answer for each question (when these are not known in advance), the difficulty levels of the questions, and the ability of each participant. We show how the model can: determine a probability distribution over answers for a given question by aggregating the responses of participants based on their abilities and the questions' difficulties; test the ability levels of participants efficiently by finding the best next question to ask in an *adaptive* way, depending on the previous responses; automatically calibrate aptitude tests from a set of questions and the responses provided by the participants, determining the relative difficulty levels of the questions and their ability to discriminate between participants of similar but uneven skill levels.

## 2. Related Work

Measuring intelligence is a key topic in psychology. Psychologists showed that peoples' performance on many cognitive tasks is strongly correlated, so a single statistical factor called "general intelligence" emerges (Anastasi et al., 1997). A measure for the performance of *groups* of people in joint tasks, called "collective intelligence", was investigated (Woolley et al., 2010). This approach focuses on explicit collaboration and interaction between members of the crowd. Although in our setting participants do not interact directly, one can view our model as a method for inferring correct answers to questions given the responses of a crowd of individuals. The number of correct answers inferred can serve as a measure of the intelligence of the crowd. In this sense, our work is somewhat similar to other approaches which also use aggregated responses to IQ tests for measuring collective intelligence (Lyle, 2008; Bachrach et al., 2012; Kosinski et al., 2012).

Psychometricians developed a body of theory called "test theory" which analyzes outcomes of psychological testing, such as the ability levels of participants or the difficulty of questions in a test, trying to improve reliability in such tests (Anastasi et al., 1997). One paradigm for designing tests of mental abilities, is the "item-response theory" (Hambleton et al., 1991) (IRT for short). IRT has been used to develop high-stakes adaptive tests such as the Graduate Management Admission Test (GMAT). IRT is based on the idea that the probability of a participant providing the correct response to a question is a function of both a parameter of the question and a parameter of the item (for example, the question's difficulty and the person's aptitude). When applying aptitude tests, the parameter of the person is *latent* (cannot be directly observed), and only its manifestation, in the form of the participant's responses, can be directly observed. Our framework relies on a probabilistic graphical model (Koller & Friedman, 2009), using themes similar to IRT.

Many papers deal with merging opinions, ranging from information aggregation in the semantic web (Kasneci et al., 2010) to prediction markets (Pennock & Sami, 2007). Frameworks such as Probabilistic Relational Models (Getoor et al., 2007) combine a logical representation with probabilistic semantics, and allow inference to aggregate information and opinions. One basic method for collective decision making is voting. Voting was studied in social choice theory (Sen, 1986), which focuses on how participants can manipulate by lying about their preferences. We assume that the experts' responses are their true opinion and focus on the inference problem. One application of our model is aggregating crowdsourced opinions. A machine learning approach for doing so which does not model task difficulty was proposed in (Raykar et al., 2010) and a technique that models task difficulty but uses an EM approach was proposed in (Whitehill et al., 2009). Another method based on graphical models is (Welinder et al., 2010). In that model questions are endowed with features, which could represent concepts or topics and participants have different areas of expertise matching these topics. Our model focuses on a general domain in the spirit of test theory and IRT, and does not rely on specific features. An active learning approach for labeling data was proposed in (Yan et al., 2011) and is similar to our adaptive IQ testing technique. Another approach akin to ours is the TrueSkill system (Herbrich et al., 2007), which uses a graphical model to estimate the relative skill levels of people based on past contests.

## 3. Joint Probabilistic Model of Difficulty, Ability, and Response

We present a probabilistic model for analyzing multiple problem domains, which we refer to as the Difficulty-Ability-REsponse estimation model, or DARE for short. The inputs to the model are responses that participants give to multiple choice questions. Additional inputs may be ground truth information for some or all of the questions. The model falls into the framework of probabilistic graphical models. Such models allow structurally describing the generative process assumed to underlie the observed data in terms of latent and observed random variables. In the domain of interest, information such as the correct response to a question, the ability of a participant, and the difficulty of a question are modeled as unobserved variables whereas the given response to a question by a user is viewed as an observed variable. The structure of the model is determined by the conditional independence assumptions made about the variables in the model. Pearl (Pearl, 1988) introduced Bayesian Networks (directed graphical models), which encode assumptions of conditional independence as a graph whose vertices represent variables and whose edges represent dependencies between variables. We use the more general notion of a factor graph, see e.g. (Koller & Friedman, 2009), to describe the factorial structure of the assumed joint probability distribution among the variables. After defining the structure of the model as a factor graph and setting the observed variables to their observed values, approximate message passing algorithms (Koller & Friedman, 2009) can infer marginal probability distributions of unknown variables of interest such as the correct response to a question, the ability of a participant, or the difficulty of a question.

### 3.1. The DARE Model

We model a situation in which a set $P$ of participants is available to answer a set $Q$ of multiple choice questions. We assume that for each question $q \in Q$ there are $R_q$ possible answers, only one of which, $y_q \in R_q$, is correct. We model the process by which participants $p \in P$ produce responses $r_{pq} \in R_q$ to questions $q \in Q$. We assume that: a) Every participant has an underlying ability $a_p \in \mathbb{R}$ which determines her ability to determine the correct answer to questions $q \in Q$. b) Each question $q$ has an inherent difficulty $d_q \in \mathbb{R}$ which determines how likely it is that participants $p \in P$ will know the correct answer to question $q$.

We propose a joint probabilistic model whose factor graph is given in Figure 1: The model has two parts, one modeling the probability of participant $p$ knowing the correct answer to question $q$ (left of $c_{pq}$ in Figure 1), and one relating the true answer $y_q$ to question $q$ to the response $r_{pq}$ given by participant $p$ depending on them knowing the correct answer as represented by $c_{pq}$ of the answer (right of $c_{pq}$ in Figure 1). Knowledge of the correct answer, $c_{pq} \in \{T, F\}$, is modeled as an interaction of the ability $a_p \in \mathbb{R}$ of participant $p$, and the difficulty $d_q \in \mathbb{R}$ of question $q$. Specifically, it is assumed to depend on the difference $t_{pq} := a_p - d_q$ via:

$$\begin{aligned} P(c_{pq} = T | t_{pq}, \tau_q) &:= \int_{-\infty}^{\infty} \phi(\sqrt{\tau_q}(x - t_{pq}))\theta(x)\, dx \\ &= \Phi\left(\sqrt{\tau_q} t_{pq}\right). \end{aligned} \quad (1)$$

Here $\phi$ denotes the standard Gaussian density $\phi(x) := \sqrt{2\pi}^{-1} \exp(-x^2/2)$ and $\Phi$ denotes the (sigmoidal) cumulative Gaussian distribution $\Phi(t) := \int_{-\infty}^{t} \phi(x)\, dx$; $\theta(\cdot)$ denotes the step function, and the precision $\tau_q$ determines how discriminative question $q$ is. The integral representation emphasizes that the probability can be viewed as emerging from a binary process resulting from evaluating the step function $\theta$ on variable $t$ with added Gaussian noise of variance $\tau^{-1}$.

The response $r_{pq}$ is modeled as a mixture of two distributions. If participant $p$ knows the correct answer to question $q$, $c_{pq} = T$, we constrain the response $r_{pq}$ to match the correct answer, $r_{pq} = y_q$, otherwise we assume that $r_{pq}$ is sampled uniformly at random from the available answers, $r_{pq} \sim \text{DiscreteUniform}(R_q)$. Note how this mixture is expressed as a *gate* (dashed pair of boxes in Figure 1), which switches the factor connecting to $r_{pq}$ depending on the state of the variable $c_{pq}$. Gates were introduced in (Minka & Winn, 2008) as a powerful and flexible notation that simplifies factor-graph representations of mixture models. They can be

How To Grade a Test Without Knowing the Answers

used to represent context-dependent conditional independence relations, and are suited for implementations of approximate message passing inference.

In order to do inference on the model, we need to define prior distributions for the variables of interest. We assume factorizing Gaussian priors for the abilities $a_p \sim \text{Normal}(\mu_p, \sigma_p^2)$ and difficulties $d_q \sim \text{Normal}(\mu_q, \sigma_q^2)$. We choose a Gaussian prior as it lets us specify a range of plausible values based on two parameters (mean and variance) per variable, and admits a relatively simple approximate inference. The factorization assumption reflects the belief that a priori knowing the difficulty of one question would not be informative about the difficulty of another question, and similarly for the abilities of participants. We also assume factorizing discrete uniform priors for the true answers $y_q \sim \text{DiscreteUniform}(R_q)$ and for the responses $r_{pq} \sim \text{DiscreteUniform}(R_q)$ for participant-question pairs. Finally, we define factorizing Gamma priors for the precision parameters $\tau_q \sim \text{Gamma}(k, \theta)$. The Gamma prior is conveniently parameterized by a shape parameter $k$ and a scale parameter $\theta$, and is the conjugate prior for the precision parameter $\tau := \sigma^{-2}$ of the normal distribution if the mean $\mu$ is known. This choice simplifies inference by approximate message passing because the posterior also takes the functional form of the Gamma distribution.

Based on the above specification we defined a joint probability distribution $p(a_p, d_q, t_{pq}, \tau_q, c_{pq}, r_{pq}, y_q)$ for specific pairs of question $q$ and participant $p$. Assuming exchangeability of questions $q$ and participants $p$ we get a model with two plates as depicted in Figure 1, where one plate runs over participants $p \in P$ and the other over questions $q \in Q$ (plates denote a replication of the fraction of the graphical model they contain). From a generative point of view, this models a table with $|P|$ rows (one for each participant $p$) and $|Q|$ columns (one for each question $q$), where entries are the responses $r_{pq}$ of participants to questions.

### 3.2. Probabilistic Inference

We show how the model can infer quantities of interest. Generally, the data is given in the form of two incomplete sets: A set of $m$ participant-question-response triples $\mathbf{R} := \{r_{p_1 q_1}, \ldots, r_{p_m q_m}\}$ and a set of $n$ ground-truth question-answer pairs $\mathbf{y} := \{y_{q_1}, \ldots, y_{q_n}\}$. One special case is when all the ground-truth question-answer pairs are known. This is the traditional test scenario as used in aptitude tests including school tests, GMAT, IQ tests, etc. Another special case is crowdsourcing domains where we may not have any ground-truth available, but can obtain participant-question-response triples depending on budget and time constraints. As shown in Section 4, providing some ground-truth question-answer pairs (a "gold-set"), can improve the accuracy of the inferred answers $y_q$ because it can be used to assess the abilities $a_p$ of the participants $p$ more accurately, leading to more accurate inference on the answers $y_q$. Generally, for every observed response $r_{pq}^*$, we set the discrete prior distribution $p(r_{pq})$ to a single point distribution concentrated on the observed response $r_{pq}^*$, and similarly for every known ground-truth question-answer pair $y_q^*$.

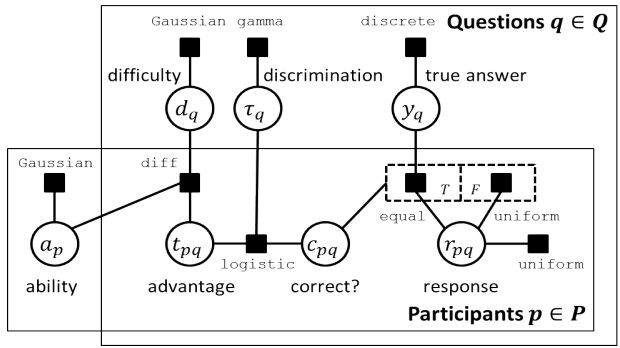

Figure 1. Factor graph for the joint difficulty-ability-response estimation (DARE) model.

Given the data $\mathbf{R}$ and $\mathbf{y}$, we wish to infer several approximate marginal (posterior) distributions: the discrete distribution $p(y_q|\mathbf{R}, \mathbf{y})$ over correct answers $y_q$, which assign a probability $\pi_q r \in [0, 1]$ to each of the possible responses $r \in R_q$; the Gaussian density $p(a_p|\mathbf{R}, \mathbf{y})$ over abilities $a_p$ of participants $p$ with means $\tilde{\mu}_p$ and variances $\tilde{\sigma}_p^2$; the Gaussian density $p(d_q|\mathbf{R}, \mathbf{y})$ over difficulties $d_q$ of questions $q$ with means $\tilde{\mu}_q$ and variances $\tilde{\sigma}_q^2$; the Bernoulli distribution $p(c_{pq}|\mathbf{R}, \mathbf{y})$ over correctness $c_{pq}$ of participant $p$'s response to question $q$ given by probabilities $\pi_{pq}$; the discrete distribution $p(r_{pq}|\mathbf{R}, \mathbf{y})$ over responses $r_{pq}$ of participant $p$ to question $q$, which assign a probability $\pi_{pqr} \in [0, 1]$ to each of the possible responses $r \in R_q$; the Gamma distribution $p(\tau_q|\mathbf{R}, \mathbf{y})$ over the precision/discrimination parameter $\tau_q$, with scale parameters $\theta_q$ and shape parameters $k_q$.

Inference in the model is done using approximate message passing (see (Koller & Friedman, 2009)). We used Infer.NET (Minka et al., 2010), a package for probabilistic inference. Specifically, we used the expectation-propagation (EP) algorithm presented in (Minka, 2001). EP allows us to calculate marginal distributions of interest on a given factor graph by iteratively calculating messages along edges that propagate information across the factor graph. In our case, EP provides only an approximation to the exact solu-tion because a) the underlying factor graph is loopy, and b) the messages at the junction between $c_{pq}$, $t_{pq}$, and $\tau_q$ are approximations, and so are the messages going in and out of the gate connected to $c_{pq}$. Thus EP is run iteratively until convergence, so its running time is linear in the input size (variables and observations).

### 3.3. Active Learning and Adaptive Testing

Having a joint probabilistic model of the data has a number of advantages, including the ability to query different distributions of interest and to handle missing data in a principled way. In addition, maintaining information about the uncertainty present in the model allows us to reason about the impact of future observations on the model uncertainty. This idea forms the basis of *active learning*, a variant of which is known in the psychometrics literature as *adaptive testing*.

Often there is a considerable cost associated with obtaining additional data points, so one can use the model of the data to determine which measurements to take next so as to improve the inferred knowledge according to a pre-determined criterion. In the absence of problem specific information, a reasonable goal is reducing uncertainty in estimates of model parameters as measured by the entropy of the posterior distributions, an idea put forward in (MacKay, 1992).

Suppose we have determined a set of model parameters of interest, denoted here as a vector $\mathbf{w}$. Our goal is to find a criterion by which to decide which response $r_{pq}$ to elicit in order to maximally reduce the posterior entropy $S(\mathbf{w})$ of those parameters defined as:

$$S(\mathbf{w}) = \int \cdots \int p(\mathbf{w}) \log \frac{m(\mathbf{w})}{p(\mathbf{w})} d\mathbf{w},$$

where $m(\mathbf{w})$ is an arbitrary base measure which does not influence the outcome (see (MacKay, 1992)). We consider two posterior distributions, $p_m(\mathbf{w}) := p(\mathbf{w}|\mathbf{R},\mathbf{y})$ before inclusion of the new data point, and $p_{m+1}(\mathbf{w}) := p(\mathbf{w}|\mathbf{R},\mathbf{y},r_{pq})$ after inclusion of data point $r_{pq}$. We then aim at maximizing the entropy reduction $\Delta S(r_{pq}) := S(p_m(\mathbf{w})) - S(p_{m+1}(\mathbf{w}))$ over the choice of response $r_{pq}$ to elicit. Since the actual response $r_{pq}$ is unknown, this choice can only be guided by the expected entropy reduction $E_{p_m(r_{pq}|\mathbf{R},\mathbf{y})}[\Delta S(r_{pq})]$, where the expectation is taken over the predictions of the model before inclusion of the new data point, i.e., based on the predictive distribution $p_m(r_{pq}|\mathbf{R},\mathbf{y})$ obtained by message passing.

In its full generality, this active learning scheme can guide the full observation/measurement process including all possible responses $r_{pq}$ and ground truth answers $y_q$. However, here we focus on the case of adaptive testing, where all the ground-truth answers $y_q$ are available, and where the goal is to determine the ability of a participant $p$ as accurately as possible, using as few questions as possible. In this special case, the parameter vector $\mathbf{w}$ only includes the ability $a_p$ of participant $p$. The posterior distribution $p_m(a_p)$ before inclusion of the new observation is Normal, $p_m(a_p) := \text{Normal}(a_p; \mu_{p.m}, \sigma^2_{p,m})$, and so is the posterior distribution after inclusion, $p_{m+1}(a_p|r_{pq}) := \text{Normal}(a_p; \mu_{p,m+1}(r_{pq}), \sigma^2_{p,m+1}(r_{pq}))$. The entropy of a univariate Gaussian with parameters $\mu$ and $\sigma^2$ is $\frac{1}{2}\ln(2\pi e \sigma^2)$, so the entropy reduction $\Delta S(r_{pq})$ is:

$$\Delta S(r_{pq}) = \frac{1}{2}\ln(\sigma^2_{p,m}/\sigma^2_{p,m+1}(r_{pq}))$$

Thus the response minimizing posterior variance is preferred. Given participant $p$, for each possible question $q$ the expectation $E_{p_m(r_{pq}|\mathbf{R},\mathbf{y})}[\Delta S(r_{pq})]$ is calculated by examining the following quantities for all possible responses $r_{pq} \in R_q$: a) their probabilities $\pi_{pq}$, and b) the resulting posterior variances $\sigma^2_{p,m+1}(r_{pq})$ in the updated model. From these we compute the expected entropy reduction for each question $q$:

$$\frac{1}{2} \sum_{r_{pq} \in R} \pi_{pq} \ln(\sigma^2_{p,m}/\sigma^2_{p,m+1}(r_{pq}))$$

We then pick the question $q^*$ that reduces the expected entropy the most.

## 4. Empirical Analysis

We empirically tested the DARE model discussed in Section 3.1 using a dataset of responses to a standard intelligence test, called Raven's Standard Progressive Matrices (SPM) (Raven), which falls within the category of multiple choice domains. It consists of sixty questions, each of which consists of a matrix of shapes with one element missing and eight possible answers. Each answer is a possible shape that completes the matrix, but only one answer is correct. A sample item, similar[1] to those in SPM is shown in Figure 2. SPM is one of the most popular intelligence tests, and was used both for research and clinical purposes.

The sample consisted of 120 individuals who filled SPM for its standardization in the British market in 2006 (Raven). The mean number of correct responses, called "raw score", was 99.57 (STD=14.16).

### 4.1. Unobserved Correct Answers

First, we investigate the DARE model's ability to handle missing correct answers $y_q$. In this case the model allows us to compute the probability $p(y_q|\mathbf{R}, \mathbf{y})$ that a given answer $y_q$ is correct. To minimize the probability of error we select the mode of that distribution as the model's answer for that question. When provided with the responses of all 120 participants the DARE model correctly infers the correct responses for 46 of the questions. Note, that the number of errors is not too surprising because some items in the test were very difficult and few participants answered them correctly. The highest raw score was fifty so even the top scoring participant answered ten items incorrectly.

---

[1] The SPM test is copyright protected, so we only provide an example question similar to those in the real test.

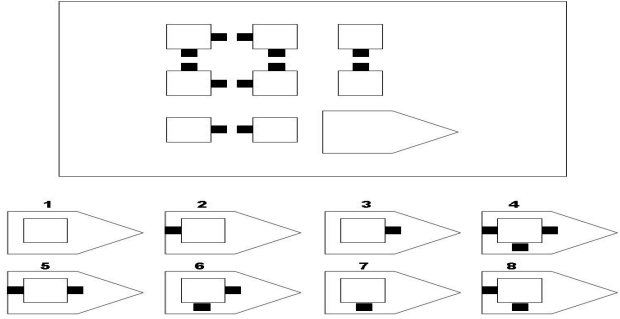

Figure 2. Item similar to those in the SPM test

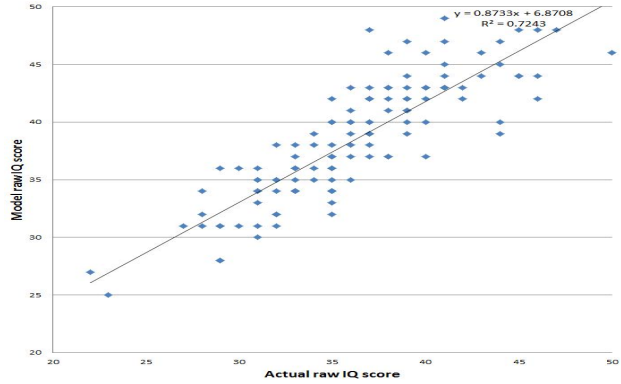

Figure 3. Estimates of skill levels for missing information regarding the correct answers to the questions.

We can calculate a participant's raw IQ score with respect to the true correct answers $y_q^*$ or with respect to the predicted "correct" answers $\hat{y}_q$, and we refer to the latter score as the *model raw score*. In crowdsourcing situations when the correct answer for each question is unknown, one can use the model raw scores as an estimate of participants' abilities. Figure 3 shows a scatter plot, in which each point represents a participant; the position on the x-axis represents the participant's raw IQ score, and the position along the y-axis their model raw score. As Figure 3 indicates, there is a very strong correlation between the true raw IQ scores and model raw scores ($R^2 = 0.7243$), and the difference between the two scores is rather small across all participants.

One can think of DARE as an aggregator that receives the responses of a crowd of participants, and outputs the inferred answer for each question. Existing work (Lyle, 2008; Bachrach et al., 2012) tests simpler aggregators using IQ test data. The former uses majority voting, and the latter does consider the ability levels of participants but assumes all items to be of equal difficulty. Another possible simplifying assumption, not examined in this earlier work, is that all the participants have equal ability. In contrast, in the DARE model, the probability of a participant to know the correct answer depends both on the difficulty $d_q$ of the question and the ability $a_p$ of the participant, with the above scenarios as special cases.

We refer to the model with different question difficulties as the *question model*, and the model with different participant abilities as the *participant model*. We examine how such simplifications affect the model's ability to infer correct answers as a function of the amount of available data. Figure 4 shows how well the question, participant and DARE models perform in this regard. For any given crowd size, shown on the x-axis, we randomly selected 10,000 subsets of participants of that size. For each such crowd we inferred the correct answer $\hat{y}_q$ to each question using the model, and used the number of questions for which the inferred answer $\hat{y}_q$ was equal to the true correct answer $y_q^*$ as a measure of the model's performance. The y-axis is the quality of each model, averaged over the 10,000 sampled crowds. Figure 4 shows the ability of all models to infer correct answers increases with the amount of data. It also shows that DARE outperforms the simpler models. Interestingly, only modeling participants ability of is better than only modeling question difficulty (which is equivalent to majority vote).

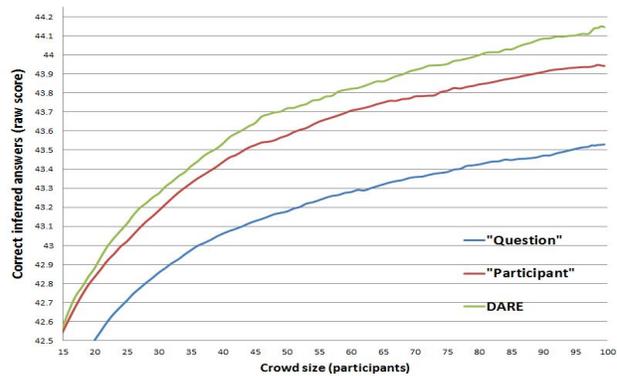

Figure 4. Effect of crowd size on correct responses inferred.

**Crowdsourcing Data:** To examine the applicability of our model to crowdsourcing, we tested our model on the TREC 2011 Crowdsourcing Track dataset (Lease & Kazai, 2011), generated by crowdsourced workers which classified search engine responses for queries (relevant / irrelevant). Each query-document pair is a "question as workers must determine if the document is relevant for the query. This dataset is sparse, as most workers only examined few "questions, and all "questions have at most 10 answers in total, and includes ground-truth judgements. We isolated the 369 "questions with the most answers (8 per "question), and the 84 workers who answered the most "questions (at least 30 answers each). Our analysis shows that DARE slightly outperforms majority voting on this dataset. Majority voting gets 206 questions correct, while DARE gets 210 correct. We also tested how well DARE estimate participants skills, similarly to Figure 3. Although for this crowdsourcing dataset the resulting scatter plot is quite noisy ($r^2$ of 0.79), it is similar to the one in Figure 3.

### 4.2. Partial Information on Correct Answers

We now examine situations where participants are first tested on a "gold-set" of questions for which the correct answer is known. Consider choosing $i$ questions and making the correct answer to these questions observable to the model. This does not reveal the correct answer to the remaining $|Q| - i$ questions, but it does allow the model to better estimate the ability levels of the participants, which in turn allows the model to better infer the correct answer to these remaining items. Figure 5 shows this effect in DARE. The x-axis represents the number of "revealed" questions and the y-axis represents the proportion of the *remaining* questions for which the model inferred the right answer. For each number $i$ of "revealed" items, we sampled $100,000$ crowds of 20 participants and $i$ revealed questions (uniformly at random), and the location on the y-axis is the average proportion of the *remaining* questions for which the model inferred the right answer over this sample. As the figure shows, having a larger "gold-set" increases the model's ability to infer the correct response for the remaining items.

### 4.3. Adaptive Skill Testing

We now show how DARE can be used for adaptive skill testing. Given a budget of $b$ questions to ask, our goal is to infer the participants' ability levels. We use DARE to estimate a participant's raw IQ score after only observing this participant's responses to a set of "asked" questions (revealed responses). In a static approach, for each budget $b$ we choose a specific question set used for *all* the participants, and measure the RMSE in estimated raw IQ across all participants. To choose the best set of questions of size $b$ for the static approach, one must enumerate over all possible $\binom{|Q|}{b}$ question sets of size $b$, and use the one minimizing the error. This is intractable when $|Q|$ is large, so we heuristically choose the question set. We selected static question set for a given budget $b$ by choosing questions that equally partition the participant population in terms of the fraction of participants who solved the question[2] For example, with a budget $b = 2$ we selected a question that roughly two thirds of the participants solved correctly and one that roughly one third of the participants solved incorrectly.

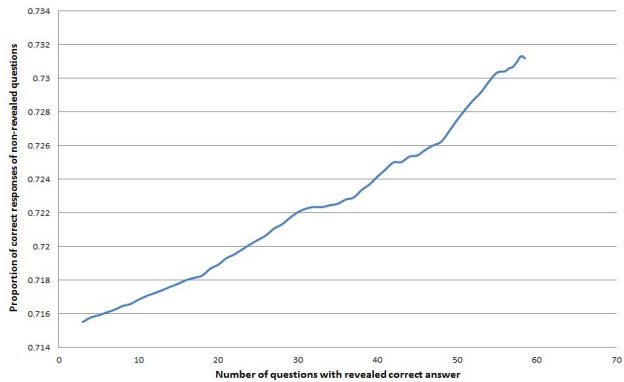

*Figure 5.* Effect of partial information on correct answers.

We also implemented the adaptive testing scheme of Section 3.3 and compared it to the baseline static approach. Under the adaptive approach, the next question to ask depends on the participant's response to earlier questions, so we reveal the participant's responses one at a time. To measure the RMSE for a given budget $b$, we simulated the adaptive process for each of the participants and averaged the errors across all participants. Figure 6 shows RMSEs for the static and adaptive approaches for different budget levels. It shows the adaptive approach has a smaller error in its inferred ability levels for any given budget. [3]

### 5. Conclusions and Limitations

We presented the DARE model for inferring the correct answers, difficulty levels of questions and abil-ity levels of participants in multiple problem domains. Our evaluation of the model shows that joint inference of these quantities is possible to a high level of accuracy and that it is indeed possible to grade a test without knowing the answers. We showed that in our setting modeling participants' ability levels is more important than questions' difficulty levels, that including a "gold-set" helps, and that active learning leads to more efficient testing.

---

[2]The static approach essentially has access to information regarding the difficulty of the questions which is not normally be available. As our analysis shows, our active approach beats the static approach even when the static approach can use such information.

[3]The standard deviation for the RMSEs is 1.07 for the adaptive scheme and 0.99 for the static scheme.

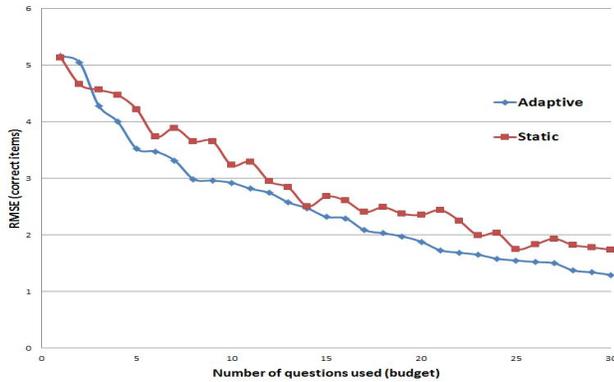

*Figure 6.* Static and adaptive skill testing.

Our approach is subject to several limitations. Our evaluation used an IQ dataset, whereas crowdsourcing tasks may exhibit different properties, such as a greater homogeneity in task difficulty levels. Also, we assume that participants answer to the best of their ability, but participants may be selfish agents with varying motives. For a game theoretic treatment of such issues see (DiPalantino & Vojnovic, 2009; Gao et al., 2012).

Many questions are open for future research. Are there better models for aggregating responses, or models better tailored to other domains? How can one tractably compute the optimal *non-adaptive* test for a given population? Can we use similar models to infer the ability levels of individuals when only their performance within the context of a *group* is known?